\documentclass[runningheads]{llncs}
\usepackage{graphicx}
\usepackage{comment}
\usepackage{amsmath,amssymb} % define this before the line numbering.
\usepackage{color}
\usepackage{multirow}

\usepackage[width=122mm,left=12mm,paperwidth=146mm,height=193mm,top=12mm,paperheight=217mm]{geometry}

\begin{document}
% \linenumbers
%\pagestyle{headings}
%\mainmatter
%\def\ECCVSubNumber{3778}  % Insert your submission number here

\title{Interlayer and Intralayer Scale Aggregation for Scale-invariant Crowd Counting} % Replace with your title

%\begin{comment}
%\titlerunning{ECCV-20 submission ID \ECCVSubNumber} 
%\authorrunning{ECCV-20 submission ID \ECCVSubNumber} 
\author{Mingjie Wang$^{1,2}$, Hao Cai$^{1,2}$, Jun Zhou$^{3}$, Minglun Gong$^{2}$}

\institute{$^{1}$ Memorial University of Newfoundland, NL, Canada\\
	$^{2}$University of Guelph, ON, Canada\\
	$^{3}$Dalian Maritime University, Dalian, China}
%\end{comment}
%******************

\begin{comment}
\titlerunning{Abbreviated paper title}
\author{First Author\inst{1}\orcidID{0000-1111-2222-3333} \and
Second Author\inst{2,3}\orcidID{1111-2222-3333-4444} \and
Third Author\inst{3}\orcidID{2222--3333-4444-5555}}
%
\authorrunning{F. Author et al.}
\institute{Princeton University, Princeton NJ 08544, USA \and
Springer Heidelberg, Tiergartenstr. 17, 69121 Heidelberg, Germany
\email{lncs@springer.com}\\
\url{http://www.springer.com/gp/computer-science/lncs} \and
ABC Institute, Rupert-Karls-University Heidelberg, Heidelberg, Germany\\
\email{\{abc,lncs\}@uni-heidelberg.de}}
\end{comment}
%******************
\maketitle

\begin{abstract}
Crowd counting is an important vision task, which faces challenges on continuous scale variation within a given scene and huge density shift both within and across images. These challenges are typically addressed using multi-column structures in existing methods. However, such an approach does not provide consistent improvement and transferability due to limited ability in capturing multi-scale features, sensitiveness to large density shift, and difficulty in training multi-branch models. To overcome these limitations, a Single-column Scale-invariant Network (ScSiNet) is presented in this paper, which extracts sophisticated scale-invariant features via the combination of interlayer multi-scale integration and a novel intralayer scale-invariant transformation (SiT). Furthermore, in order to enlarge the diversity of densities, a randomly integrated loss is presented for training our single-branch method. Extensive experiments on public datasets demonstrate that the proposed method consistently outperforms state-of-the-art approaches in counting accuracy and achieves remarkable transferability and scale-invariant property.
\end{abstract}

\section{Introduction}
\label{introduction}

Research in crowd counting has gained popularity in recent years due to its important real-world applications, such as video surveillance, traffic control, public security, and scene understanding~\cite{guerrero2015extremely,onoro2016towards,zhang2017understanding,zhang2017fcn,hsieh2017drone}.
With the unprecedented success of Convolutional Neural Networks (CNNs) in computer vision tasks, a series of  CNN-based crowd counting algorithms have been developed in the past 6 years~\cite{idrees2013multi,li2013anomaly,zhang2016single,sam2017switching,sindagi2017generating,liu2018leveraging,ranjan2018iterative,jiang2019crowd,li2018csrnet,liu2019context,liu2019crowd,sindagi2019multi}.  While these works have shown to be effective, accurate counting under challenging scenarios (e.g. severe occlusions, perspective effects, background clutter, scale variation, and density shift~\cite{xu2019learn}) is still not yet achieved. 

\begin{figure}[h]
	\centering
		\includegraphics[height=6.5cm,width=0.9\linewidth]{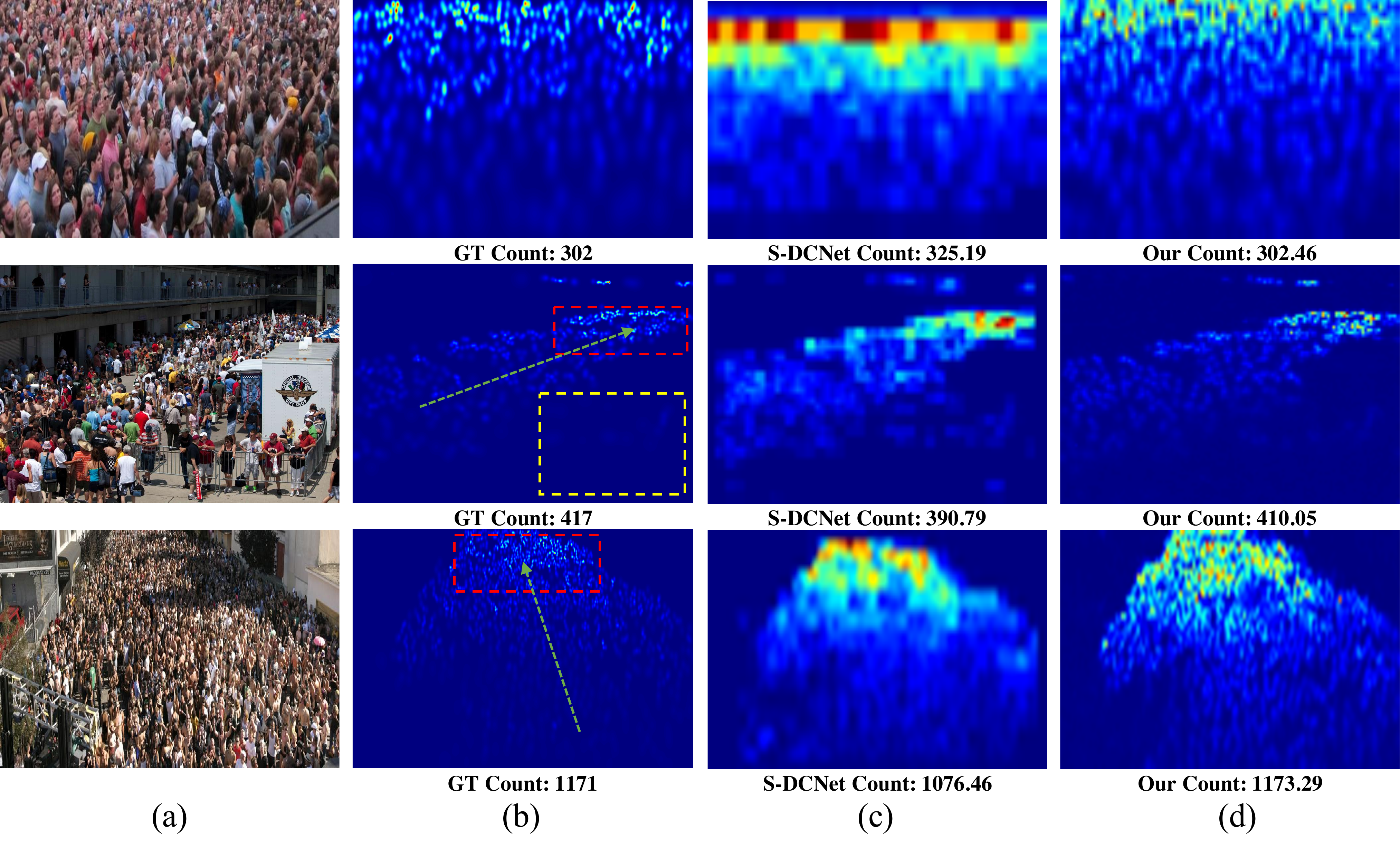}
	\caption{Samples (a) taken from the ShanghaiTech Part\_A dataset~\cite{zhang2016single} demonstrate the challenges of continuous scale changes within a single image and high density shifts both within and across images. Green arrows in the ground truth (GT) density maps (b) show the directions of scale changes, whereas red and yellow rectangular boxes highlight high and low density areas, respectively. Compared to S-DCNet~\cite{xiong2019open} (c), our method (d) handles these challenges much more effectively.} 
	\label{fig:scalevariation}
\end{figure}

Previous studies have shown that scale variation in image to some extent limits the performance of designed models in vision tasks like object detection~\cite{Lin_2017_CVPR,cai2016unified,najibi2017ssh} and semantic segmentation~\cite{zhao2017pyramid,chen2016attention,roy2016multi}. This issue is even more pronounced in the field of crowd counting as the scale in those crowd images tends to vary continuously and dramatically.  Fig.~\ref{fig:scalevariation}(a) shows examples of scale variation within a single image and density shift across crowded scenes.

Considerable efforts have been made to deal with this issue by either utilizing multi-column CNNs~\cite{zhang2016single,liu2019crowd,sam2017switching} or a stack of multi-branch units~\cite{cao2018scale} to extract features at multiple scales, or using skip connections to fuse feature maps from different layers by leveraging the fact that deeper layers produce higher-level information with larger reception fields~\cite{sindagi2019multi,shi2019revisiting,liu2019context,jiang2019crowd} whereas shallower layers encode low-level spatial information for accurate localization. Nevertheless, these efforts largely lie on layer-wise multi-scale feature extraction in a coarse-grained fashion which takes the whole layer as input, resulting in insufficient diversity of scales. This inevitably affects the robustness and performance of their models. For instance, the S-DCNet~\cite{xiong2019open}, which has the lowest Mean Absolute Error (MAE) to date, does not generate accurate counts; see Fig.~\ref{fig:scalevariation}(c).

Recent work~\cite{wu2018group} has suggested that different channels of visual features are not entirely independent and they somewhat resemble traditional features like SIFT~\cite{lowe2004distinctive} and HOG~\cite{dalal2005histograms}, where each group of channels is constructed by some type of histogram.
Motivated by this observation, we here propose to integrate both interlayer and intralayer scale fusion for scale-invariant crowd counting. To achieve intralayer scale fusion, different groups of feature channels are used to encode different regions of an image with varying receptive fields, allowing fine-grained multi-scale features to be captured. Combining this novel intralayer scale aggregation with the conventional interlayer multi-scale feature extraction allows the presented method to learn more sophisticated and diverse scale information, leading to more accurate performance; see Fig.~\ref{fig:scalevariation}(d).

\begin{figure}[h]
	\begin{center}
		\includegraphics[height=5.5cm,width=\linewidth]{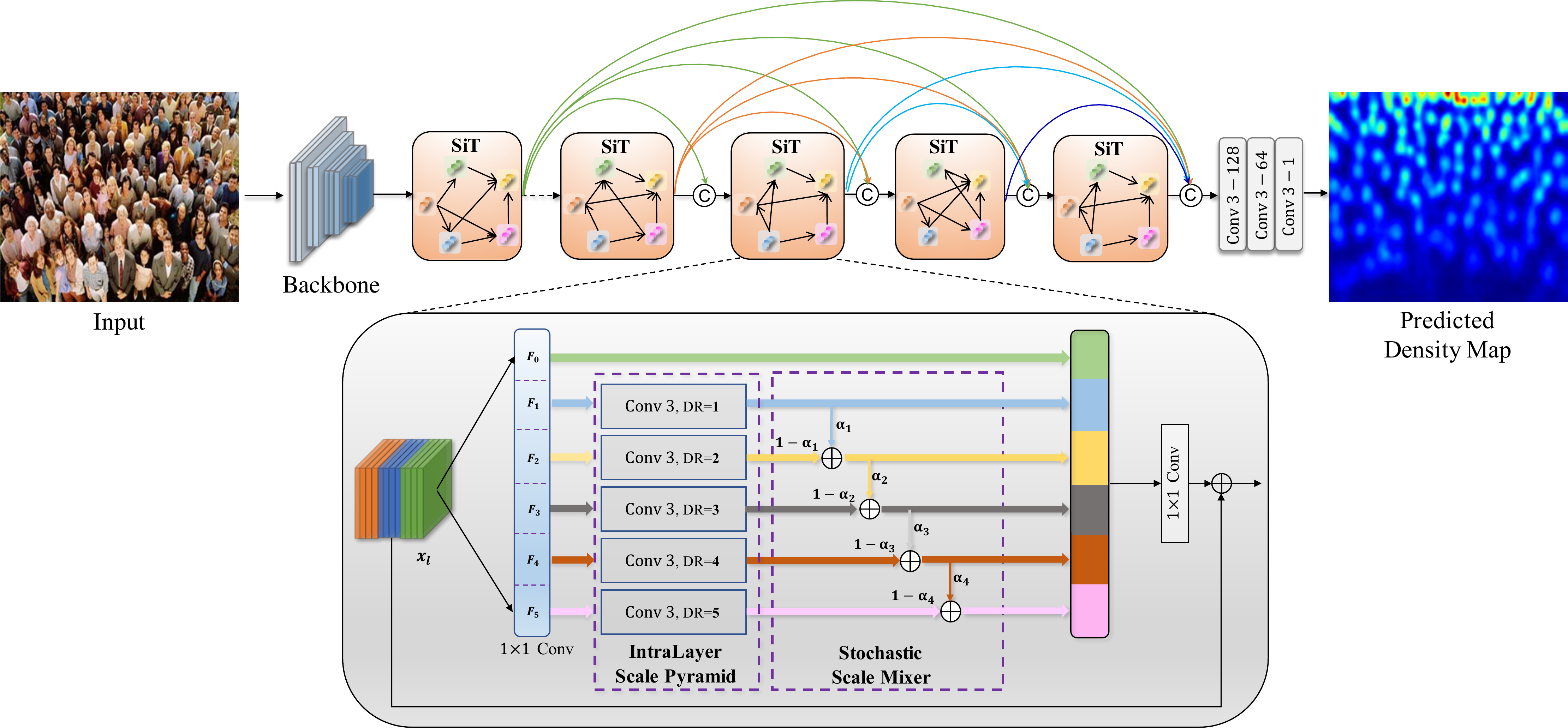}
	\end{center}
	\caption{Pipeline of the proposed ScSiNet for crowd counting. The backbone subnetwork is formed by the first ten layers of a pre-trained VGG-16~\cite{simonyan2014very} to produce general features. The SiTs are combined by interlayer dense connections~\cite{huang2017densely} and each SiT is comprised of an intralayer scale pyramid and a stochastic scale mixer. Specifically, the input of SiT is first processed by an 1$\times$1 layer and then divided into $G=6$ groups, denoted as $F_0,...,F_5$. The scale pyramid performs dilated convolutions for groups $F_1,...,F_5$ with dilation rates (DR) ranging from 1 to 5. This is followed by the scale mixer, which is designed to blend the multi-scale features in a stochastic fashion. $\alpha_1,\alpha_2,\alpha_3,\alpha_4$ are random variables following the uniform distribution between 0 and 1.} 
	\label{fig:model}
\end{figure}

Specifically, this paper presents a novel Single-column Scale-invariant Network (ScSiNet) architecture for crowd counting. As shown in Fig.~\ref{fig:model}, the proposed ScSiNet consists of a VGG-based backbone subnetwork, six scale-invariant transformation (SiT) layers with dense connections~\cite{huang2017densely} in between, and output layers. The pattern of dense connections is leveraged to perform interlayer fusion of multi-level features  through forwarding detailed features from shallower layers to all subsequent high-level layers. The SiT conducts fine-grained intralayer scale fusion and is the key component in our ScSiNet. Experimental results in Sec.~\ref{experiments} indicate that this novel SiT contributes to the robustness and stabilization of our model.

Furthermore, to overcome the grand effects of large density shifts and small training samples in crowd counting datasets like UCF\_CC\_50~\cite{idrees2013multi}, we introduce a simple yet effective objective function, which we call \emph{randomly integrated loss}. Specifically, during each iteration, a batch of training sample patches with consistent resolution are randomly cropped and then processed separately by the model; see Sec.~\ref{minibatch}. Finally, an \emph{integrated loss} is computed to optimize the network. 
In this way, the limited training sets can be better made use of and thus provides our model with resistance to overfitting. Moreover, the stochastic batch of patches with varying densities and integrated loss mitigate the density fluctuations, thereby enabling our model to be robust to huge density shifts without using any multi-column structure.

The effectiveness of the proposed model is evaluated and compared against existing
state-of-the-art methods on three widely-used datasets: ShanghaiTech~\cite{zhang2016single}, UCF-QNRF~\cite{idrees2018composition} and UCF\_CC\_50~\cite{idrees2013multi}. Our contributions are summarized as follows:
\begin{itemize}
\item A novel intralayer scale-invariant transformation (SiT) module is designed, which effectively solves the problem of continuous scale variation in congested scenes and avoids potential overfitting.
\item A randomly integrated loss is proposed, which significantly mitigates the effects of huge density shift in crowd datasets.
\item The proposed ScSiNet outperforms the state-of-the-art methods and exhibits superior transferability and scale-invariance property, which are highly desired in real-world applications.
\end{itemize}
\section{Related Work}

Numerous algorithms are proposed in the past decade for counting objects in congested scenes. Traditional methods~\cite{ge2009marked,dollar2011pedestrian,viola2005detecting} primarily focus on counting-by-detection that explicitly detects individual objects being counted. Since the performance of detection is often affected by occlusion and background clutter, which in turn limits the accuracy of these counting-by-detection approaches. Later, the regression-based counting methods~\cite{chen2012feature,chen2013cumulative,lempitsky2010learning} improve the counting accuracy through learning a mapping between image features and total count/density map. 

More recently, CNN-based crowd counting methods~\cite{idrees2013multi,li2013anomaly,zhang2016single,sam2017switching,sindagi2017generating,liu2018leveraging,ranjan2018iterative,jiang2019crowd,li2018csrnet,liu2019context,liu2019crowd,sindagi2019multi} have achieved dramatic improvements via the efforts in designing various creative architectures to handle the issues of scale and density variations. Notable works include MCNN~\cite{zhang2016single}, Switching-CNN~\cite{sam2017switching}, HydraCNN~\cite{onoro2016towards}, AMDCN~\cite{deb2018aggregated} and DSSINet~\cite{liu2019crowd}. These models aim at learning multi-branch regressors to extract and fuse multi-scale features, which improves the robustness with respect to scale variation. Meanwhile, a number of methods focus on other build techniques such as: $i)$ bottom-top and top-bottom fusion of features from different layers~\cite{sindagi2019multi}; $ii)$ pursuing structural consistency in predicted density maps using auxiliary loss functions, such as composition loss~\cite{idrees2018composition}, adversarial loss~\cite{shen2018crowd}, SSIM~\cite{cao2018scale,liu2019crowd}, and Bayesian Loss~\cite{ma2019bayesian}; $iii)$ the use of dilation and deformable convolutions to enlarge the receptive fields~\cite{li2018csrnet,deb2018aggregated,guo2019dadnet,liu2019adcrowdnet}; $iv)$ leveraging unlabeled data via a self-learning strategy~\cite{liu2018leveraging} and negative correlation-based learning~\cite{shi2018crowd}; and $v)$ utilizing attention mechanisms to adaptively aggregate multi-scale features~\cite{liu2019context,cao2018scale,sindagi2019multi,zhang2019attentional}. 
More recently, efforts have been devoted to refining the subregions of predicted density maps. For instance, Cheng \emph{et al.}~\cite{cheng2019learning} propose a new Maximum Excess over Pixels (MEP) loss to learn the spatial awareness, whereas Xu \emph{et al.}~\cite{xu2019learn} adopt multipolar center loss to bring the selected regions to multiple similar scale levels.
Zhang \emph{et al.}~\cite{zhang2019attentional} and Liu \emph{et al.}~\cite{liu2019crowd} propose to utilize conditional random fields to fuse complementary features.  

Although the aforementioned CNN-based methods have made excellent progress in tackling scale variation, they all concentrate on layer-wise transformations, postprocessing, or ad-hoc loss terms, resulting in merely coarse-grained scale information. In comparison, we propose to perform intralayer, fine-grained scale fusion through assigning different groups of feature channels with receptive fields of different sizes. Integrating both interlayer and intralayer scale aggregation enhances our model with the capability of encoding multi-scale information from subregions of feature maps.
Additionally, current crowd counting algorithms usually suffer from severe overfitting due to over-complex fusions or multi-branch structures, leading to poor generalization capability. To this end, Liu \emph{et al}.~\cite{liu2018leveraging} introduce a self-learning strategy leveraging abundant unlabeled images to alleviate overfitting, whereas Shi \emph{et al}.~\cite{shi2018crowd} build decorrelated regressors to produce generalizable features. Both approaches are shown to be effective, but the performance gain is limited. Inspired by regularization techniques applied in the task of image classification, such as mixup~\cite{zhang2017mixup} and shake-shake regularization~\cite{gastaldi2017shake}, we design a novel scale mixer to fuse fine-grained multi-scale features via progressively stochastic convex combinations, which regularizes the scale dimension and further boosts the transferability and scale invariance of our model.

\section{Proposed Method}

%In this section, we first discuss how interlayer and intralayer scale aggregations are performed, before introducing the proposed uniform mini-batch training strategy.

\subsection{Interlayer Dense Connections}

It is well known that different layers in deep neural networks correspond to different receptive fields and semantic levels. Generally, representations from shallow layers have smaller semantic levels, and the encoded low-level spatial details can be used to obtain better localization~\cite{sindagi2019multi}. Meanwhile, high-level features extracted by deeper layers have larger scales, which facilitate the retrieval of contextual information. Intuitively, in order to exploit the merits of multi-level features for crowd counting, it is important to integrate features at different semantic levels/scales. As shown in Fig.~\ref{fig:model} (top), we fuse layer-based multi-scale information through dense connections~\cite{huang2017densely}. Specifically, for the $l^{th}$ transformation, the input $x_l$ at multiple layer-based scales can be formulated as $x_{l}=[y_0,y_1,...,y_{l-1}]$,
%\begin{equation}
%	x_{l}=[y_0,y_1,...,y_{l-1}],
%	\label{equ:denseconnection}
%\end{equation}
where $[y_0,y_1,...,y_{l-1}]$ denotes the concatenation of multi-scale features from preceding layer $0,1,...,l-1$. 

As mentioned in Sec.~\ref{introduction}, current CNN-based models of crowd counting rely heavily on \emph{interlayer} multi-scale feature extraction to deal with the large scale variation, most of which utilizing multi-column CNNs or a stack of multi-branch units. The reuse of multi-scale features through dense connections enables the proposed network achieve comparable performance with a single-column architecture, which simplifies training and prevents overfitting. In addition, we also exploit \emph{intralayer} multi-scale fusion to seek more diverse representation of scales at different granular levels, as discussed in the subsection below. 

\subsection{Intralayer Scale-invariant Transformation}
\label{FSIT}

Recent research~\cite{wu2018group} has demonstrated that feature vectors in deep CNNs are not unstructured vectors and hence it is reasonable to group channels and expect different groups to capture different visual cues, such as shapes, frequency, and textures. Successful applications of group-based convolutions have been found in the task of image classification~\cite{xie2017aggregated,chollet2017xception,zhang2018shufflenet}. Motivated by this, we believe that features in a given layer could be grouped in terms of image sub-regions with different scales. To this end, we design a novel intralayer scale-invariant transformation (SiT), which involves an intralayer scale pyramid and a stochastic scale mixer as shown in Fig.~\ref{fig:model} (bottom). The scale pyramid has the benefit of guiding the network to learn multi-scale representations corresponding to independent sub-regions without increasing computational overhead.
Given that traditional aggregation methods such as weighted summation or concatenation have limited abilities to approximate continuous scales and are prone to overfitting, the scale mixer is proposed to shuffle and regularize fine-grained multi-scale features through a progressively stochastic manner. The scale mixer enables the network with two-fold advantages: 
$i)$ decorrelating the groups of scale pyramid to regularize the features at different scales; 
$ii)$ reducing sensitivity to scale changes to some extent.

\subsubsection{Intralayer Scale Pyramid}

Instead of capturing features using groups of standard $3\times3$ filters in other group-based models~\cite{xie2017aggregated,chollet2017xception,zhang2018shufflenet}, we split a feature vector into $G$ groups, where each group is empirically set to contain 64 channels \cite{he2016deep}. Due to dense connections, the number of channels in the input $x_l$ varies at different layer $l$. To convert different numbers of input channels to $G$ groups of $64$ channels, an $1\times 1$ convolution layer is used.  The output of the $1\times 1$ convolution layer is denoted as $[F_0, F_1, F_2,...,F_{G-1}]$, where each $F_i$ has 64 channels. 

To guide different groups to capture varied scales, dilated convolutions with different dilation rates are used to build a scale pyramid. Dilated convolution has the advantage of enlarging receptive field sizes without increasing the number of parameters~\cite{li2018csrnet}. Specifically, here $F_0$ is fed directly to the output (without passing through convolutions) as it aims to preserve the high-frequency information. The remaining $G-1$ groups go through dilated convolutions with dilation rates from 1 to $G-1$, respectively, to capture features at corresponding scales.  The obtained intralayer scale pyramid can be written as $[F_0,$ $D_1(W_1,F_1),$ $D_2(W_2,F_2),...,D_{G-1}(W_{G-1},F_{G-1})]$,
%\begin{equation*}
%[F_0,D_1(W_1,F_1),D_2(W_2,F_2),...,D_{G-1}(W_{G-1},F_{G-1})],
%\label{equ:scalepyramid}
%\end{equation*}
where $[,]$ denotes the concatenation of groups with different receptive fields and $W_i$ represents the trainable parameters of dilation convolutions $D_i$ with dilation rate $i$, $i \in [1,2,...,G-1]$. Examples can be found in Fig.~\ref{fig:model} (bottom), where it is divided into 6 groups and delivers $1\times1$ $(F_0)$, $3\times3$ $(D_1)$, ... ,$11\times 11$ $(D_5)$ receptive fields respectively.

\subsubsection{Stochastic Scale Mixer}
%\begin{figure}[t]
%	\begin{center}
%		\includegraphics[width=\linewidth]{scalesblender.pdf}
%	\end{center}
%	\caption{The illustration of discrete scale pyramid and continuous range of scales produced by stochastic scale blender. \mlc{this figure is not convincing since it does not relate to any real data...}} 
%	\label{fig:blender}
%\end{figure}
In the scale pyramid, each convolution only operates on the corresponding group of channels, which brings several side effects: $i)$ The output from a certain group only relates to the input within the group~\cite{zhang2018shufflenet} and this blocks cross-group information communication; $ii)$ The scale variation in crowd image is usually continuous and scale change  boundaries are often fuzzy, whereas the pyramid can only generate strict boundaries of scales which could result in oversensitivity of the network.

To tackle these problems, a novel scale mixer is designed to progressively regularize and blend the multi-scale representations through a stochastic tree structure. For simplicity, we denote the outputs of scale pyramid and scale mixer as $[d_0, d_1, d_2,...,d_{G-1}]$, $[\hat d_0, \hat d_1, \hat d_2,...,\hat d_{G-1}]$, respectively. The procedure of scale mixer can be represented as a recursive function:
\begin{equation}
\left\{
\begin{array}{lr}  
\hat d_i=d_i,  i=0,1 &\\
\hat d_i=\alpha_{i-1}\times \hat d_{i-1}+ (1-\alpha_{i-1})\times d_i,  i=2,...,G-1, &
\end{array}
\right.
\label{equ:scalesblender}
\end{equation}
where $\alpha_i$ are random variables following the uniform distribution between 0 and 1. Before each iteration in training, all variables are overwritten with new random values, while they are set to expected value of 0.5 at the test time. Taking $G=6$ as an example, we deduce $\hat d_5$ as follows:
\begin{equation}
\begin{split}
\hat d_5=&\alpha_4\hat{d_4}+(1-\alpha_4)d_5=\alpha_4[\alpha_3\hat{d_3}+(1-\alpha_3)d_4]+(1-\alpha_4)d_5 =\alpha_1\alpha_2\alpha_3\alpha_4d_1\\&+(1-\alpha_1)\alpha_2\alpha_3\alpha_4d_2+(1-\alpha_2)\alpha_3\alpha_4d_3 + (1-\alpha_3)\alpha_4d_4 + (1-\alpha_4)d_5.
\end{split}
\label{equ:blenderexample}
\end{equation}
Note that the mixer complements cross-scale features through a set of random variables, thus aiding in approximating the continuous scale change in crowd images and blur scale change boundaries.
Moreover, the stochastic scale mixer is differentiable, which means it can be embedded into network for end-to-end training. 
%The Jacobian of the aggregation $\hat d_5$, for example, can be computed by:
%{\small
%\begin{equation}
%\begin{split}
%\frac{\nabla\hat d_5}{\nabla x}&=\alpha_1\alpha_2\alpha_3\alpha_4\frac{\nabla d_1}{\nabla x} + (1-\alpha_1)\alpha_2\alpha_3\alpha_4\frac{\nabla d_2}{\nabla x}\\ +(1&-\alpha_2)\alpha_3\alpha_4\frac{\nabla d_3}{\nabla x} +
%(1-\alpha_3)\alpha_4\frac{\nabla d_4}{\nabla x} + %(1-\alpha_4)\frac{\nabla d_5}{\nabla x}. 
%\end{split}
%\label{equ:gradients}
%\end{equation}
%}
Depending on the random variables, the scale mixer can smoothly vary its behaviour among multiple scales, thus helping produce more sophisticated and diverse scale information. Finally, feature maps transformed by the scale mixer are concatenated to be passed through an $1\times1$ convolutional layer to reduce the width of filters to 256 and assemble information together. To make our structure easier to be optimized, the paradigm of residual learning~\cite{he2016deep} is incorporated into our proposed SiT.

\subsection{Randomly Integrated Loss} \label{minibatch}

\begin{figure}
	\centering
	\includegraphics[width=\linewidth,height=4cm]{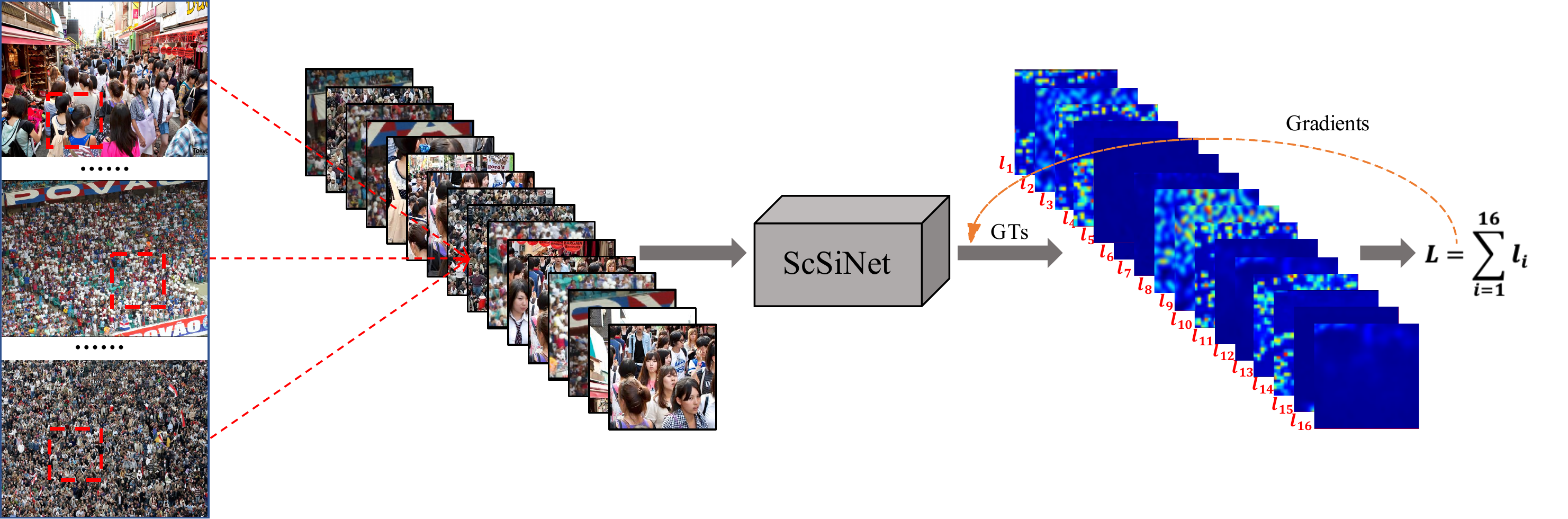}
	\caption{Schema of randomly integrated loss with patch number $N=16$. During each iteration, $N$ patches randomly cropped from various original images are fed into our ScSiNet and finally an integrated loss is computed as the supervision.}
	\label{fig:joint}
\end{figure}
%It is well known that mini-batch training has become a standard component in other tasks of computer vision like image classification and semantic segmentation due to its merits and interested readers are referred to~\cite{ioffe2015batch}. However, the advantages of this technique are not deeply exploited in most existing works of crowd counting and the training strategies for counting networks

The current network training strategies for CNN-based counting models are mainly categorized into two classes: $i)$ Original images or 1/4 resolution patches are forwarded through the model one by one (batch size is 1) to optimize the network, where the training is dramatically affected by the number of samples and density shifts across the datasets; $ii)$ A batch of patches randomly cropped in an off-line fashion are fed into the network and an average value of losses is obtained as the supervision. For instance, given the batch size $N$, the traditional loss~\cite{liu2019context,li2018csrnet,zhang2016single} is computed as follows:

{\small
\begin{equation}
\begin{split}
L=\frac{1}{2N}\sum_{i=1}^{N}||Y_i-GT_i||_2^2 = \sum_{i=1}^{N}||\frac{1}{\sqrt{2N}}Y_i-\frac{1}{\sqrt{2N}}GT_i||_2^2,
\label{equ:averageloss}
\end{split}
\end{equation}}
where $Y_i$ is the predicted density map for $i^{th}$ image and $GT_i$ means corresponding ground truth. 

As it can be seen in Eq.~\ref{equ:averageloss}, both $Y_i$ and $GT_i$ are divided by $\sqrt{2N}$ and their densities follow normal distribution  $\mathcal{N}(0, \sigma^2)$ with variance $\sigma$ indicating the range of densities in crowd datasets. As a result, this loss function inevitably changes the initial distributions and reduces the density diversity especially when the patch size is small or the batch size $N$ is large. 
In this case, inappropriate cropping size and batch size easily lead to the difficulty of convergence or trapping in local optimal solutions. Although several attempts~\cite{ma2019bayesian,xu2019learn,xiong2019open,liu2019context,idrees2018composition,liu2019crowd} have been made to overcome this issue by setting different sizes for each dataset or ad-hoc loss terms to remedy the the degradation of density diversity, these approaches often cause inconsistent improvements and degraded transferability across the datasets.

In this work, we propose a simple yet effective randomly integrated loss to better address the grand effects of large density shifts and limited training samples at widely-spread resolutions in crowd datasets.
That is, during each iteration, we randomly crop patches online from original images with a constant resolution for all datasets and form the input with the batch size of $N$; see Fig.~\ref{fig:joint}.
Instead of using an average value of losses as in Eq.~\ref{equ:averageloss}, we obtain the batch loss by summing all sub-losses $l_1,l_2,...,l_N$, and the objective function is defined as: 
\begin{equation}
L=l_1+l_2+...+l_N=\sum_{i=1}^{N}||Y_i(x_i;W)-GT_i||_2^2,
\label{equ:loss}
\end{equation}
where $W$ denotes the trainable parameters of the network, $Y_i(x_i;W)$ means the corresponding estimated density map patch and $GT_i$ is the ground truth of $x_i$. 

By feeding in randomly cropped patches, it allows our single-column model to possess the merits of ``divide-and-conquer''  without resorting to the extremely complicated patch-based subnetworks used by recent approaches~\cite{xiong2019open,xu2019learn}.
Additionally, the limited training samples can be fully utilized, while the density range is further enlarged and closely continuous, thereby enabling our single-column network to be easily trained and robust to huge density shifts.

%The problem may caused by the strategy is inconsistent distributions of internal representations. Thus it is necessary to normalize internal features of different patches to accelerate the network training, which can be easily implemented by using batch normalization~\cite{ioffe2015batch} with the following formulation:
%\begin{equation}
%{Norm(d^i_j)}=\gamma^i\frac{\hat{d^i_j}-E[\hat{d^i}]}{\sqrt{Var[\hat{d^i}]}}+\beta^i, ~~j\in [1,2...N]
%\label{equ:batchnormalization}
%\end{equation}
%where $E[\hat{d^i}]$ and $Var[\hat{d^i}]$ are the pixel-wise mean and variance values of joint patches $\hat d^i$,  $\gamma^i$ and $\beta^i$ are a pair of learnable parameters to scale and shift the normalized values. 
\section{Experimental Results}
\label{experiments}

The following three widely used datasets are employed to evaluate the performance of our proposed method:

\noindent
{\bf ShanghaiTech.} The ShanghaiTech dataset contains 1,198 crowd images with a total of 330,165 annotated people and is divided into two parts. {\bf Part\_A} consists of 482 scenes with 300 images for training and the rest for testing. {\bf Part\_B} includes 716 images where 400 images are for training.

\noindent
{\bf UCF-QNRF.} This is the latest large-scale crowd dataset, containing 1,535 high resolution images with annotated people ranging from 49 to 12,865, thus making this dataset feature huge density variance. The training and testing sets consist of 1,201 and 334 images, respectively.

\noindent
{\bf UCF\_CC\_50.} The UCF\_CC\_50 dataset includes 50 annotated crow images with crowd counts ranging from 94 to 4,543. This is a more challenging dataset due to the extremely limited number of samples. Following the standard protocol~\cite{idrees2013multi}, we use 5-fold cross-validation to evaluate the performance of our proposed model on this dataset. 

\subsection{Implementation Details}

%\paragraph{Ground Truth Generation.} 
Following the previous work~\cite{li2018csrnet}, we use geometry-adaptive kernels with $\beta=0.3$ and $k=3$ to blur the annotated images for ShanghaiTech Part\_A, whereas for Part\_B, UCF-QNRF, and UCF\_CC\_50, we blur all head annotations using fixed Gaussian kernels with $\sigma=15$ to generate ground-truth density maps.
%\paragraph{Network Training.} 
Our single-column structure is trained in an end-to-end fashion. We use Adam optimizer with a learning rate of 0.0001 to minimize the loss of Eq.\ref{equ:loss}. At each training iteration, 16 image patches with a size of $176\times 176$ are randomly cropped from original images. The first ten layers of the pre-trained VGG-16 are deployed to initialize the corresponding layers and the rest trainable weights are initialized by a normal distribution with zero mean and standard deviation of 0.01. We also add random horizontal flip for data augmentation following~\cite{sindagi2019multi}. The proposed method is implemented with Pytorch~\cite{paszke2017automatic}. Following previous work in this
area~\cite{liu2019crowd}, Mean absolute error (MAE) and mean square error (MSE) are also used as metrics to evaluate our model.
%\paragraph{Evaluation Metrics.} 
%Mean absolute error (MAE) and mean square error (MSE) are two metrics widely used to evaluate the performance of crowd counting algorithms~\cite{liu2019crowd}. They are defined as  follows:
%{\small
%\begin{equation}
%\begin{split}
%MAE=\frac{\sum^{\bar{N}}_{i=1}|C_i-C_i^{GT}|}{\bar{N}},
%MSE=\sqrt{\frac{\sum^{\bar{N}}_{i=1}|C_i-C_i^{GT}|^2}{\bar{N}}},
%\end{split}
%\label{equ:metrics}
%\end{equation}
%}
%where $\bar{N}$ is the total number of test images, $C_i$ and $C_i^{GT}$ are the predicted and target counts for the $i^{th}$ sample, which are obtained by integrating the density maps.

\begin{table}
	\begin{center}
		\caption{
			Comparison with state-of-the-art methods on ShanghaiTech (Part\_A and Part\_B), UCF-QNRF and UCF\_CC\_50 datasets. Best results are shown in boldface.
		}
		\label{table:comparision}
		\begin{tabular}{c|cc|cc|cc|cc}
			\hline
			\multirow{2}{*}{Methods}& \multicolumn{2}{c|}{{\bf Part\_A}} 
			&\multicolumn{2}{c|}{{\bf Part\_B}}&\multicolumn{2}{c|}{{\bf UCF-QNRF}}&\multicolumn{2}{c}{{\bf UCF\_CC\_50}}\\
			\cline{2-9}
			~& MAE & MSE & MAE & MSE  & MAE & MSE  & MAE & MSE\\
			\hline
			TEDNet~\cite{jiang2019crowd} & 64.2 & 109.1 & 8.2 & 12.8  & 113 & 188 & 249.4 & 354.5\\
			ADCrowdNet~\cite{liu2019adcrowdnet}& 63.2 & 98.9 & 7.6 & 13.9  & - & - & 257.1 & 363.5\\
			PACNN + CSRNet~\cite{shi2019revisiting} & 62.4 & 102.0 & 7.6 & 11.8  & - & - & 241.7 & 320.7\\
			CAN~\cite{liu2019context}  & 62.3 & 100.0 & 7.8 & 12.2 & 107 & 183 & 212.2 & 243.7\\
			
			CFF~\cite{shi2019counting} & 65.2	& 109.4 & 7.2 &	12.2 & - & - & - & -\\
			
			SPN+L2SM~\cite{xu2019learn} & 64.2 & 98.4 & 7.2 & 11.1  & 104.7 & 173.6 & 188.4 & 315.3\\
			MBTTBF-SCFB~\cite{sindagi2019multi} & 60.2 & 94.1 & 8.0 & 15.5 & 97.5 & 165.2 & 233.1 & 300.9\\
			PGCNet~\cite{yan2019perspective}& 57.0 & {\bf 86.0}   & 8.8 & 13.7 & - & - & 244.6 & 361.2\\
			BL~\cite{ma2019bayesian}  & 64.5 & 104.0  & 7.9 & 13.3 & 92.9 & 163.0 & 237.7 & 320.8\\
			DSSINet~\cite{liu2019crowd} & 60.63 &	96.04  & 6.85 &	10.34 & 99.1	&	 {\bf 159.2} & 216.9 &	302.4\\
			SPANet+SANet~\cite{cheng2019learning} & 59.4	& 92.5  & {\bf 6.5}	& {\bf 9.9} & -	&	- & 232.6 & 311.7\\
			S-DCNet~\cite{xiong2019open} & 58.3 &	95.0  & 6.7	& 10. & 104.4	&	176.1 & 204.2	& 301.3\\
			\hline
			\hline
			{\bf ScSiNet} (proposed) & {\bf 55.77} & 90.23 & 6.79 &  10.95 & {\bf 89.69} & 178.46 & {\bf 154.87} & {\bf  199.42} \\
			\hline
		\end{tabular}
	\end{center}
\end{table}
\setlength{\tabcolsep}{1.4pt}

\subsection{Comparison with the State-of-the-Arts}

Table~\ref{table:comparision} compares the proposed ScSiNet with several state-of-the-art methods across three crowd counting datasets. The results show that our ScSiNet achieves the best MAE performance on ShanghaiTech Part\_A, UCF-QNRF, and UCF\_CC\_50 datasets, even though its performance on Part\_B is slightly lower than SPANet + SANet~\cite{cheng2019learning} and S-DCNet~\cite{xiong2019open}.  It is also worth noting that both UCF-QNRF and UCF\_CC\_50 are more challenging than ShanghaiTech, as UCF-QNRF has wider density distributions and larger image resolutions whereas UCF\_CC\_50 has the highest density images and extremely small training set. On UCF-QNRF, ScSiNet produces a significant improvement of 3.21$\downarrow$ in MAE compared with the existing best approach (BL~\cite{ma2019bayesian}). On UCF\_CC\_50, ScSiNet lowers the MAE from 188.4 produced by SPN+L2SM~\cite{xu2019learn} to 154.87 ($17.8\%$ $\downarrow$), and MSE from 243.7 of CAN~\cite{liu2019context} to 199.42 ($18.2\%$ $\downarrow$).

\begin{figure}
	\centering
	\includegraphics[width=0.8\linewidth,height=8cm]{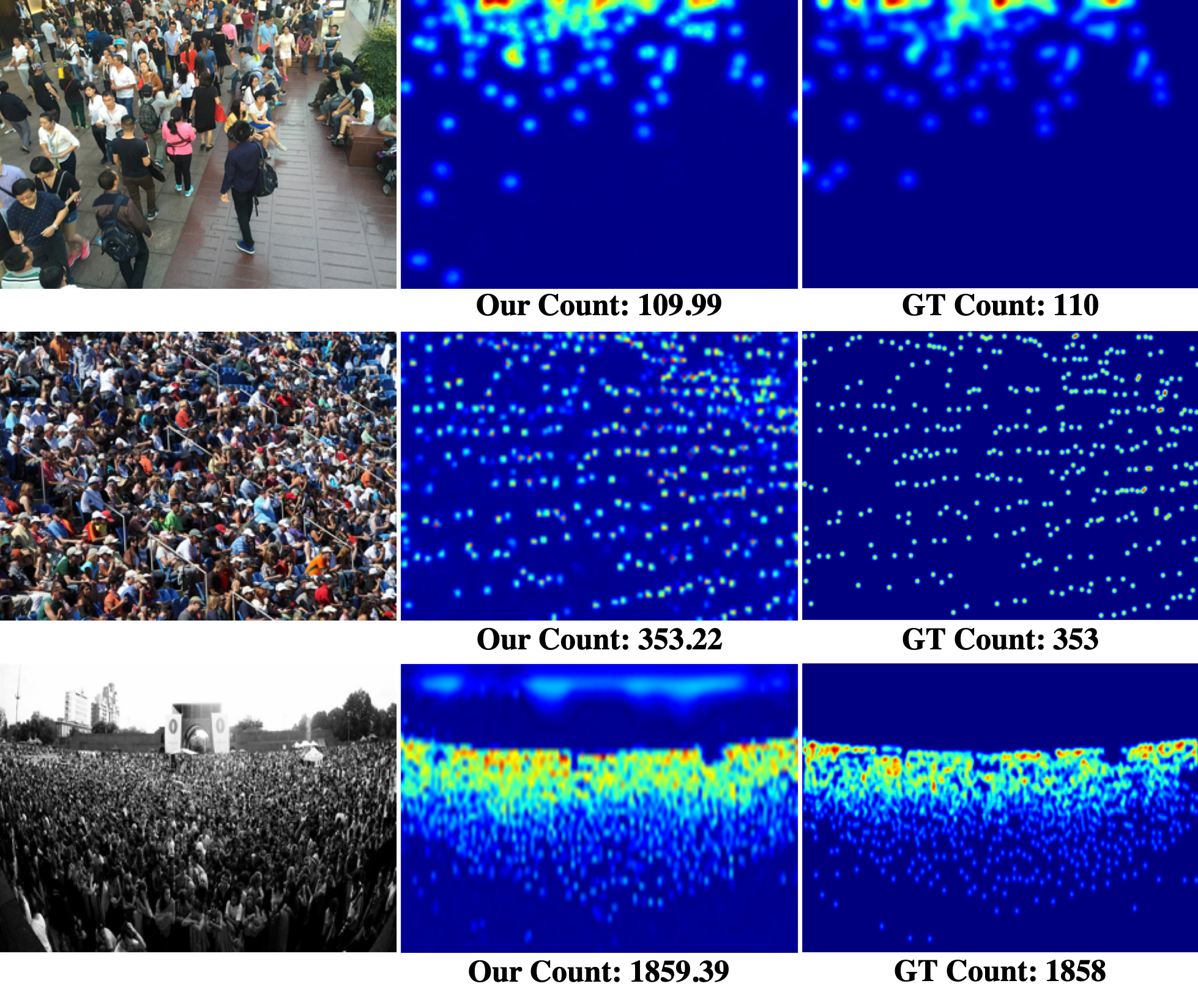}
	\caption{Visualizations of prediction examples on ShanghaiTech Part\_B (top), UCF-QNRF (middle), and UCF\_CC\_50 (bottom). %The left column shows the original images, middle column exhibits the predicted counts produced by proposed model and the right column displays the ground-truth density maps.
	} 
	\label{fig:Bexample}
\end{figure}

Qualitative comparisons with S-DCNet~\cite{xiong2019open} for three samples with different densities from the ShanghaiTech Part\_A are presented in Fig.~\ref{fig:scalevariation}. As it can be seen that the presented ScSiNet estimates more accurate count and meanwhile generates higher-quality predicted density maps (Fig.~\ref{fig:scalevariation}(d)) with less noise than the current state-of-the-art S-DCNet~\cite{xiong2019open} (Fig.~\ref{fig:scalevariation}(c)).  Samples of the test cases from ShanghaiTech Part\_B, UCF-QNRF, and UCF\_CC\_50 are demonstrated in Fig.~\ref{fig:Bexample}. The computational time are also measured by calculating the average inference time on the test set of Part\_A. Run on a single Titan Xp GPU, the average overhead of ScSiNet is about 0.072s per image, which is 64\% faster than that of S-DCNet~\cite{xiong2019open} at 0.20s per image.

%\begin{figure}[h]
%	\begin{center}
%		\includegraphics[width=\linewidth]{Aexample.pdf}
%	\end{center}
%	\caption{Qualitative visualizations of the proposed method and S-DCNet~\cite{xiong2019open} on ShanghaiTech Part\_A. From left to right: original crowd images, predicted density map given by S-DCNet, density maps generated by our method and the ground truth.} 
%	\label{fig:Aexample}
%\end{figure}

\subsection{Scale-invariant Tests}

\paragraph{Robustness to Resolution Changes.}
Most of the crowd counting models are trained and tested on high-resolution images.  This leads to high memory or computational costs and limits the generalization on devices with lower computational capacity, such as smart security cameras. Reducing the resolution of testing images changes scale distributions and hence causes performance drop for non-scale-invariant methods. 

\begin{figure}
	\centering
	\includegraphics[width=0.85\linewidth,height=5.5cm]{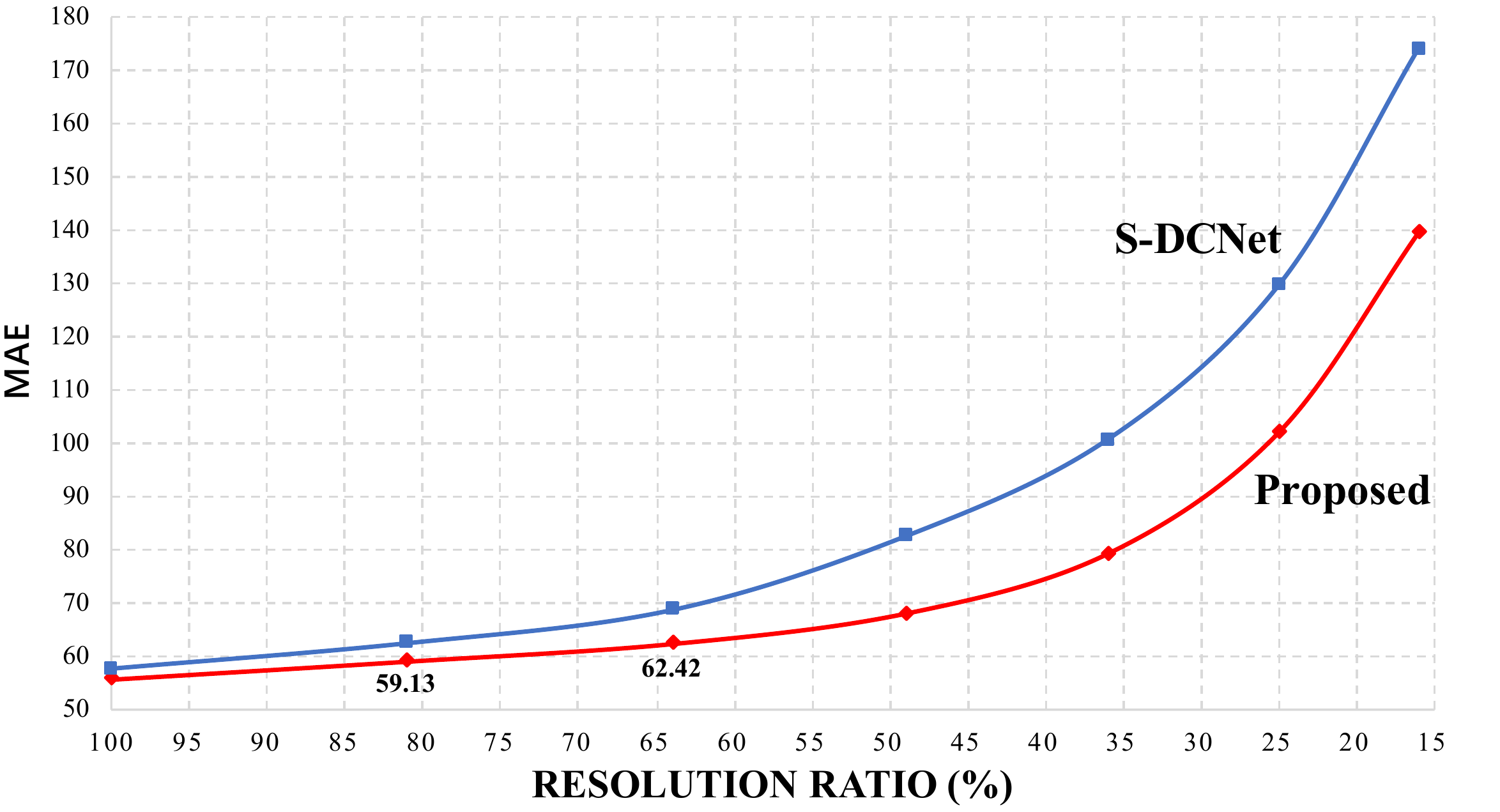}
	\caption{Performance evaluation for ScSiNet and S-DCNet~\cite{xiong2019open} on the original ShanghaiTech Part\_A dataset (100\%) and its downsampled versions. As image resolution reduces, the performance gain of ScSiNet over S-DCNet becomes more prominent.%, which demonstrates ScSiNet's superior scale-invariant property.
	} 
	\label{fig:resize}
\end{figure}

To test how well the proposed ScSiNet can handle resolution drops, we downsample the test images from ShanghaiTech Part\_A by various ratios, ranging from 100\% (unchanged) to 16\% (40\% downsample in each dimension). The performances of ScSiNet and S-DCNet~\cite{xiong2019open} on different resolution test images are plotted in Fig.~\ref{fig:resize}. As expected, the MAE increases for both algorithms as image resolution drops. However, the increase for ScSiNet (Params. 14.1M) has a much slower rate than that of S-DCNet (Params. 28.24M). This indicates that ScSiNet can learn more scale-invariant features with less computational costs.
In addition, when ScSiNet is fed with images downsampled by 81\%, it achieves competitive MAE of 59.13, lower than those of MBTTBF-SCFB (60.2)~\cite{sindagi2019multi}, DSSINet (60.63)~\cite{liu2019crowd}, and SPANet+SANet (59.4)~\cite{cheng2019learning} achieved on full-resolution images. Even using the images downsampled by 64\%, ScSiNet still outperforms BL~\cite{ma2019bayesian} (64.5), MPCL~\cite{xu2019learn} (64.2), and Focus for free~\cite{shi2019counting} (65.2).

\paragraph{Cross-dataset Transferability.}

\setlength{\tabcolsep}{4pt}
\begin{table}
	\begin{center}
		\caption{ Cross-dataset evaluation for transferability comparison among different methods on ShanghaiTech A\&B and UCF-QNRF (Q) datasets. %\mj{A and B represents Part\_A and Part\_B respectively while Q denotes UCF-QNRF}.
		}
		\label{table:transfer}
		\begin{tabular}{c|cc|cc|cc|cc}
			\hline
			\multirow{2}{*}{Methods}& \multicolumn{2}{c|}{{A$\rightarrow$B}} 
			&\multicolumn{2}{c|}{{B$\rightarrow$A}}&\multicolumn{2}{c|}{{ A$\rightarrow$Q}}&\multicolumn{2}{c}{{Q$\rightarrow$A}}\\
			\cline{2-9}
			~& MAE & MSE & MAE & MSE  & MAE & MSE  & MAE & MSE\\
			\hline
			D-ConvNet~\cite{shi2018crowd}  & 49.1 & 99.2 & 140.4 & 226.1 &-&-&-&-\\
			SPN~\cite{xu2019learn}  & 23.8 & 44.2 & 131.2 & 219.3  & 236.3 & 428.4 & 87.9 & 126.3\\
			SPN+L2SM~\cite{xu2019learn}  & 21.2 & 38.7 & 126.8 & {\bf 203.9}  & 227.2 & 405.2 & 73.4 & 119.4\\
			\hline
			\hline
			{\bf ScSiNet} & {\bf 20.96} & {\bf 36.38} & {\bf 118.19} &  214.13 & {\bf 194.63} & {\bf 370.85} & {\bf 69.23} & {\bf 107.44} \\
			\hline
		\end{tabular}
	\end{center}
\end{table}
\setlength{\tabcolsep}{1.4pt}

To illustrate the transferability of our proposed ScSiNet, we conduct cross-data experiments, where the model is trained on one specific dataset and validated on other datasets without any retraining or fine-tunning. As can be seen from Table~\ref{table:transfer}, our method consistently outperforms D-ConvNet~\cite{shi2018crowd} and SPN+L2SM~\cite{xu2019learn} by a remarkable margin. Note that the ScSiNet model trained on UCF-QNRF even outperforms some methods on ShanghaiTech Part\_A like ic-CNN~\cite{ranjan2018iterative} (MAE 69.8 and MSE 117.3) and L2R~\cite{liu2018leveraging} (MAE 73.6 and MSE 112.0) that are trained on the same dataset.
The superior transferability can be attributed to ScSiNet's scale-invariant property as well, as the network is robust against scale and density variations across different datasets.

\subsection{Ablation Study}

To further understand the effectiveness of the proposed ScSiNet and randomly integrated loss, we perform a detailed ablation study over the three datasets.

\paragraph{Effects of Inter/Intra-layer Aggregations and Optimal G value in SiT}. In order
to demonstrate the importance of both interlayer and intralayer scale fusion, we first evaluate two variants of our model: No Interlayer and No Intralayer. The variant with no interlayer fusion is achieved by removing dense connections, whereas that with no intralayer setting degenerates to the DenseNet-like model. Due to lack of group convolution, no intralayer setting introduces a larger number of parameters. The ablative results show that both interlayer
and intralayer are contributive to the improvements of counting performance; see Table~\ref{table:groupnumber}. As discussed in Sec.~\ref{FSIT}, the hyper parameter G used in intralayer scale
pyramid balances computational overhead and scale diversity. Here, we evaluate
ScSiNet with different G values in Table 3 as well. It can be observed that our
method performs well across different G values and achieves the best performance
when G=6, which corresponds to 1.4M total parameters. Higher G value results
in a modest increase in MAE due to potential overfitting under high model
complexity.

\setlength{\tabcolsep}{4pt}
\begin{table}
	\begin{center}
		\caption{The impacts of inter/intra-layer scale fusion and the number of groups (G) in SiT on ShanghaiTech Part A dataset.}
		\label{table:groupnumber}
		\begin{tabular}{c|c|c|c|c|c|c}  %{\hsize}{@{}@{\extracolsep{\fill}}ccccc@{}}
			\hline
			& No Inter (G=6) & No Intra (G=1) &G=2&G=4&G=6&G=8 \\
			\hline
			\hline
			Parameters& 11.0M & 20.9M &11.7M&12.9M&14.1M&15.3 M \\
			\hline
			MAE& 57.00 & 60.11 &60.04& 58.85 &{\bf 55.77}&57.50 \\
			MSE& 96.77 & 104.67 &104.50& 95.83 &{\bf 90.23}& 98.60 \\
			\hline
		\end{tabular}
	\end{center}
\end{table}
\setlength{\tabcolsep}{1.4pt}

\begin{figure}
	\centering
	\includegraphics[width=0.8\linewidth,height=7.5cm]{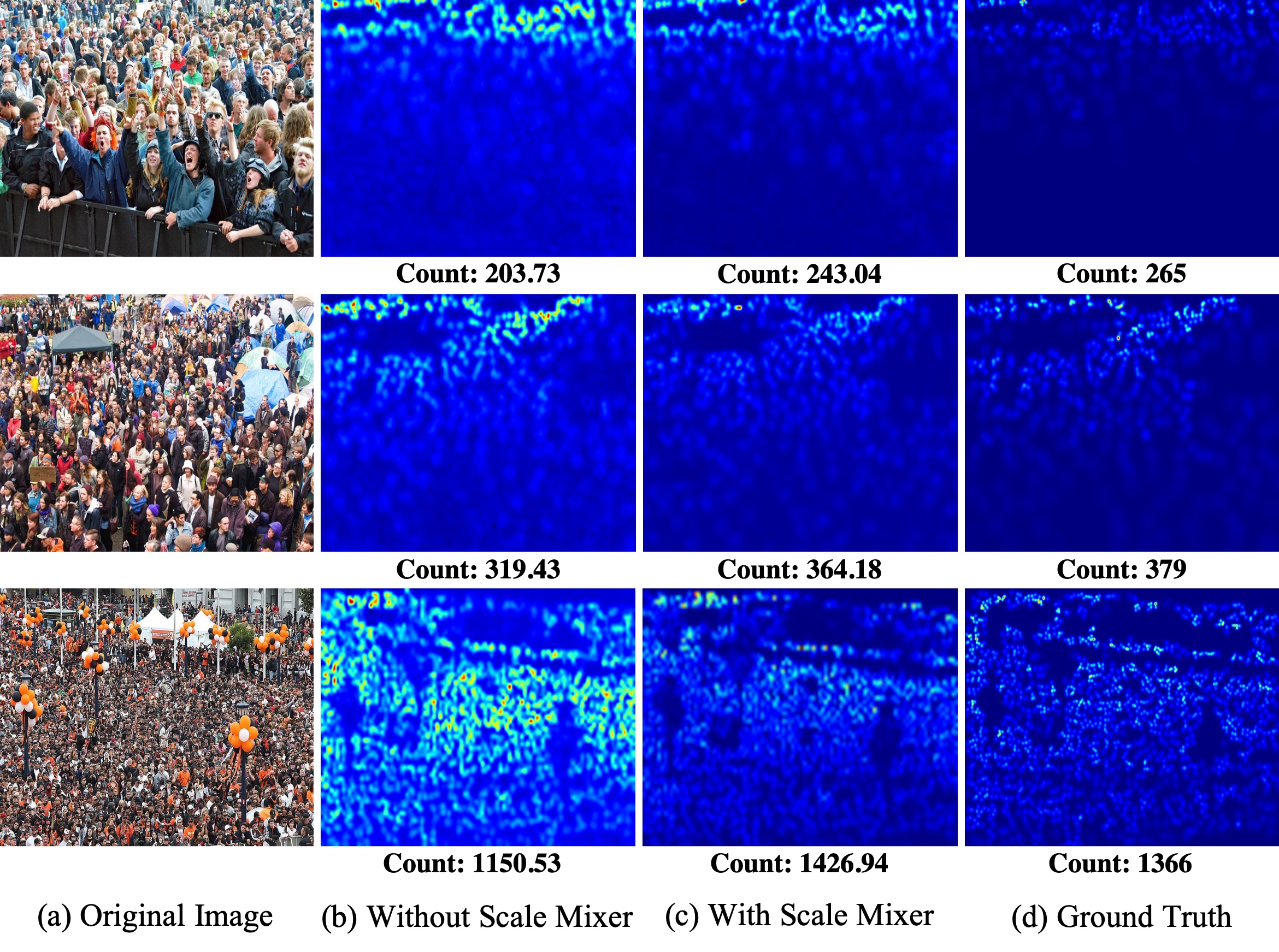}
	\caption{Qualitative evaluation of the effectiveness of stochastic scale mixer. The three representative samples are from the ShanghaiTech Part A dataset and have varied densities.} 
	\label{fig:blendersamples}
\end{figure}

\paragraph{Effectiveness of Stochastic Scale Mixer.}
We analyse the effectiveness of stochastic scale mixer on ShanghaiTech Part\_A, Part\_B and UCF-QNRF datasets. We compare the performances of the ScSiNet with and without scale mixer attached. The details are listed in Table~\ref{table:blendereffects}, which demonstrates that our proposed scale mixer indeed plays a crucial role in the counting accuracy thanks to its stronger capability of integrating features at varying scales. To better understand the importance of stochastic fusion, we also compare the effects of the scale aggregation under fixed $\alpha$=1 and stochastic $\alpha$ respectively. As it can be observed from Table~\ref{table:blendereffects}, consistent improvements are obtained from the stochastic strategy. This may be explained by that the stochastic $\alpha$ alleviates the sensibility to huge scale changes, thereby overcoming potential overfitting.
Representative samples under both settings are displayed in Fig.~\ref{fig:blendersamples}, which shows that the scale mixer is able to suppress the noisy details effectively while preserving sufficient scale information, thus helping to generate improved density maps.

\setlength{\tabcolsep}{4pt}
\begin{table}
	\begin{center}
		\caption{The effectiveness of stochastic scale mixer on ShanghaiTech Part\_A, Part\_B, and UCF-QNRF datasets. The setting (under $\alpha$=1) is not tested on UCF-QNRF due to limited computational resource.} 
		\label{table:blendereffects}
		\begin{tabular}{c|cc|cc|cc}
			\hline
			& \multicolumn{2}{c|}{w.o Scale Mixer} & \multicolumn{4}{c}{w. 
			Scale Mixer}\\
			& \multicolumn{2}{c|}{}& \multicolumn{2}{c|}{(fixed $\alpha$=1)} & \multicolumn{2}{c}{(stochastic $\alpha$)}\\
			\cline{2-7}
			& MAE & MSE & MAE & MSE & MAE & MSE\\
			\hline
			\hline
			Part\_A& 61.05 & 100.27 & 57.60 & 93.12 &55.77&90.23\\
			Part\_B& 7.01 & 11.53 & 6.96 & 11.21 &6.79&10.95\\
			UCF-QNRF& 91.12 & 170.80 & -& - & 89.69 & 178.46 \\
			\hline
		\end{tabular}
	\end{center}
\end{table}
\setlength{\tabcolsep}{1.4pt}

\paragraph{Effects of Randomly Integrated Loss.}
%\setlength{\tabcolsep}{4pt}
%\begin{table}
%	\begin{center}
%		\caption{The effectiveness of the proposed uniform mini-batch training on ShanghaiTech Part\_A and UCF\_CC\_50.}
%		\label{table:cropsize}
%		\begin{tabular}{c|cc|cc}
%		\hline
%		& \multicolumn{2}{c|}{w.o. Mini-Batch} & \multicolumn{2}{c}{w. Mini-Batch}\\
%		\cline{2-5}
%		& MAE & MSE & MAE & MSE\\
%		\hline
%		\hline
%		Part\_A& 69.2 & 116.1 & 55.77 & 90.23\\
%		UCF\_CC\_50& 229.53 & 343.27 & 154.87 & 199.42 \\
%		\hline
%		\end{tabular}
%	\end{center}
%\end{table}
%\setlength{\tabcolsep}{1.4pt}

To explore the effectiveness of the proposed randomly integrated loss,  we evaluate this strategy on Part\_A and UCF\_CC\_50. The use of the proposed loss enables the network to be robust to huge density shifts even under very small training set (MAE 69.2 $\rightarrow$ 55.77 on Part\_A and 229.53 $\rightarrow$ 154.87 on UCF\_CC\_50).  As a result, significant improvement is achieved when compared to the training using single images (batch size is 1).  The batch size $N$ is a hyperparameter in our loss, it controls the balance between the computational costs and estimation precision. We test the model trained with N=8, 16, 32, and finally N is fixed as 16 with both high accuracy and low GPU usage.

\section{Conclusion}

In this paper, we propose a Single-column Scale-invariant Network (ScSiNet) to address the issues of scale variation, density shift, and overfitting in crowd counting. We extract and integrate multi-scale features via the combination of coarse-grained interlayer dense connections and the novel intralayer scale-invariant transformation (SiT). Furthermore, a randomly integrated loss is presented to make full use of small training sets and enlarge density diversity, which enables ScSiNet to be robust against huge density shifts. Extensive experiments demonstrate that the proposed method consistently and significantly outperforms the state-of-the-art methods. The prominent transferability of ScSiNet on datasets with varied density distributions and the strong capability of ScSiNet in handling downsampled test images are also clearly illustrated.
%For future work, we plan to apply ScSiNet or its variants to other problems, such as object detection and semantic segmentation.  We also like to explore the possibility of implementing ScSiNet on mobile platforms such as smart security cameras.

\clearpage
\bibliographystyle{splncs04}
\bibliography{egbib}
\end{document}